\documentclass[letterpaper, 10 pt, conference]{ieeeconf}  %

\IEEEoverridecommandlockouts                              %

\usepackage{graphics} %
\usepackage{epsfig} %
\usepackage{mathptmx} %
\usepackage{times} %
\usepackage{amsmath} %
\usepackage{amssymb}  %
\usepackage{tabularx}
\usepackage{booktabs}
\usepackage{makecell}
\usepackage{hyperref}
\usepackage{todonotes}
\usepackage{array}

\newcommand{\mypar}[1]{\vspace{1mm}\noindent {\bf #1}~~}
\newcolumntype{C}[1]{>{\centering\arraybackslash}p{#1}}

\title{\LARGE \bf
Touch2Touch: Cross-Modal Tactile Generation for Object Manipulation
}

\author{Samanta Rodriguez*, Yiming Dou*, Miquel Oller, Andrew Owens and Nima Fazeli%
\thanks{*These authors contributed equally.  }%
\thanks{$^{1}$University of Michigan
        {\tt\footnotesize {\{samanrod, ymdou, oller, ahowens, nfz\}}@umich.edu}  }%
\thanks{Supported by NSF GRFP \#2241144 and NSF CAREER Award \#2339071, and NSF NRI \#2220876.}%
}

\begin{document}

\maketitle
\thispagestyle{empty}
\pagestyle{empty}

\begin{abstract}

Today's touch sensors come in many shapes and sizes. This has made it challenging to develop general-purpose touch processing methods since models are generally tied to one specific sensor design. We address this problem by performing cross-modal prediction between touch sensors: given the tactile signal from one sensor, we use a generative model to estimate how the same physical contact would be perceived by another sensor. This allows us to apply sensor-specific methods to the generated signal. We implement this idea by training a diffusion model to translate between the popular GelSlim and Soft Bubble sensors. As a downstream task, we perform in-hand object pose estimation using GelSlim sensors while using an algorithm that operates only on Soft Bubble signals. The dataset, the code and additional details can be found at \href{https://www.mmintlab.com/research/touch2touch/}{https://www.mmintlab.com/research/touch2touch/}.

\end{abstract}

\section{INTRODUCTION}

Tactile sensing is a fundamental enabling technology for dexterous manipulation. 
Yet, in comparison to other common modalities like vision and sound, touch sensors are much less standardized. For example, the robotics community has recently demonstrated a number of important manipulation capabilities \cite{oller2023manipulation, kim2022active, suresh2023midastouch} using a wide variety of vision-based tactile sensors, including GelSight~\cite{yuan2017gelsight}, Soft Bubble \cite{softbub_tedrake}, GelSlim \cite{gelsim_donlon}, Finger Vision \cite{fingervision}, DIGIT \cite{lambeta2020digit}, and  DenseTact \cite{Do2022DenseTactOT} to name just a few. 
This  diversity has led to a serious problem: specialized algorithms must be developed for each particular sensor. These algorithms cannot be directly used when their corresponding sensors are unavailable, and they require time-consuming modifications to adapt to other sensors. 
Likewise, machine learning models trained on one tactile sensor may not generalize to others due to large distribution shifts.

\begin{figure}[ht!]
    \centering
    \includegraphics[width=\columnwidth]{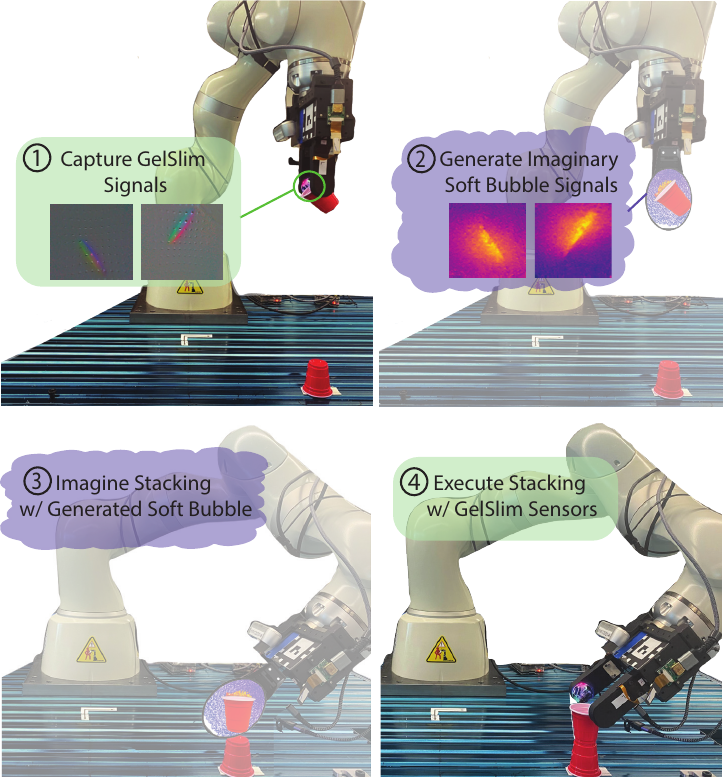}
     \caption{{\bf Transferring manipulation methods between touch sensors using cross-modal prediction.} We execute an object manipulation skill designed for one touch sensor (Soft Bubble) on a robot equipped with another sensor (GelSlim). To do this, we use a cross-modal diffusion model to translate one touch signal to another --- that is, we predict what the object would have felt like if it were manipulated with Soft Bubble rather than GelSlim. The robot then uses this prediction to perform its action.}\vspace{-0.5cm}
    \label{fig:teaser}
\end{figure}

Despite their differences, existing tactile sensors nevertheless perceive the world in similar ways. For example, vision-based touch sensors that provide representations of shape are ubiquitous~\cite{johnson2009retrographic,yuan2017gelsight,gelsim_donlon,fingervision,lambeta2020digit,Do2022DenseTactOT}. We propose using this overlapping information to convert between tactile signals obtained by different sensors, thereby enabling models designed for one touch sensor to be transferred to another. 

We formulate this problem as a cross-modal prediction task. We obtain paired touch data by having a robot touch the same objects in the same position using two different sensors. We then train a latent diffusion model~\cite{rombach2022high} to predict one touch signal from another. We demonstrate the efficacy of our method by translating from GelSlim signals to Soft Bubble signals. This translation is a challenging task because the two sensors are very different in shape, size, and compliance.

We evaluate the effectiveness of our approach on two object manipulation tasks: stacking and insertion (Fig.~\ref{fig:teaser}). A robot equipped with a GelSlim sensor manipulates an object, then its touch signals from the GelSlim are converted to Soft Bubble signals using our cross-modal prediction method. We apply a simple, off-the-shelf Iterative Closest Point (ICP) method for in-hand pose estimation designed for Soft Bubble signals to these converted touch signals. Successful downstream performance therefore requires structural properties of the touch signal to be predicted with a high degree of accuracy. Our experiments suggest: 

\begin{itemize}%[leftmargin=*]
    \item We can obtain a dataset suitable for cross-modal touch translation by having a robot automatically probe corresponding positions on an object with two touch sensors.
    \item Diffusion models can successfully estimate tactile signals captured from different sensors.
    \item Algorithms developed for one touch sensor can be transferred to others via cross-modal prediction.
    \item Insertion and stacking tasks can successfully be performed on signals obtained by cross-modal prediction.
\end{itemize}

\section{RELATED WORKS}

\mypar{Visuotactile sensors.}
In the last decade, the robotics community has adopted a variety of vision-based tactile sensors, such as GelSight~\cite{yuan2017gelsight,johnson2009retrographic}, Soft Bubble~\cite{softbub_tedrake}, GelSlim~\cite{gelsim_donlon}, Finger Vision~\cite{fingervision}, DIGIT~\cite{lambeta2020digit}, and  DenseTact~\cite{Do2022DenseTactOT}. These sensors convert touch signals into vision-like signals, representing touch as 2D images or 3D representations (e.g., point clouds). These sensors are rapidly gaining popularity and have proven valuable in a variety of applications~\cite{calandra2018more,digit_tactile_sensor,oller2023manipulation,li2014localization,kim2022active}. %
Here, we use the Soft Bubble and GelSlim visuotactile sensors in our experiments. The Soft Bubble is composed of a thin, highly compliant, air-filled membrane paired with a camera-based depth sensor. Tactile signatures are perceived as deformations of the membrane due to external contacts. The GelSlim measures deformations of an elastomeric skin using an RGB camera. %
The elastomer's opaque contact interface is illuminated by multi-colored LEDs, and the changes in color conveys deformation. We choose these two sensors because of their vastly different deformations and compliance, contact areas, image quality, and 3D (vs. 2D) representation. %

\mypar{Tasks and algorithms for tactile sensors.}
Many existing manipulation perception and controls representations are tied to specific touch sensors. For example, a variety of methods leverage sensor specific local geometry, contact force estimation, or texture~\cite{yang2023generating,li2014localization}. %
Further, in-hand object pose estimation algorithms have been developed for different visuotactile sensors (e.g., for Soft Bubble~\cite{kuppuswamy2019fast}, GelSlim~\cite{kim2022active}, and DIGIT~\cite{suresh2023midastouch}), for local geometry estimation (e.g., Soft Bubble~\cite{kuppuswamy2019fast}, GelSlim~\cite{taylor2022gelslim}, and DIGIT~\cite{xu2023visual}), for force field estimation across the contact area (e.g., Soft Bubble~\cite{kuppuswamy2020soft}, GelSlim~\cite{taylor2022gelslim}, Finger Vision~\cite{yamaguchi2016combining}). 
These methods have improved success on manipulation tasks, such as peg-in-hole insertion~\cite{kim2022active}, drawing and in-hand pivoting~\cite{oller2023manipulation}, and dense packing~\cite{ai2024robopack}. We reduce the need for sensor-specific methods by enabling models to transform one touch signal to another. 

\mypar{Cross-modal generation.}
A variety of early generative models transformed images from one format to another~\cite{isola2017image, sangkloy2017scribbler, hu2017toward, wang2018high}.
Recent works in cross-modal image translation frequently use diffusion~\cite{sohl2015deep} for its ability to generate high-quality images with stable training. These models have been used with a variety of different conditioning signals, resulting in models that perform text-to-image~\cite{avrahami2022blended,kawar2023imagic,nichol2021glide,saharia2022photorealistic,rombach2022high,zhang2023adding,ruiz2023dreambooth}, audio-to-image~\cite{girdhar2023imagebind}, video-to-audio~\cite{luo2024diff}, etc. %
Our work is closely related to methods that estimate touch from vision. These works have proposed models under various settings, including desktop~\cite{li2019connecting}, object-centric~\cite{gao2023objectfolder}, sub-scene~\cite{yang2022touch,yang2024binding} and full-scene~\cite{dou2024tactile}. Like many of these works~\cite{higuera2023learning,yang2023generating,yang2024binding,dou2024tactile, caddeo2024sim2real}, we use diffusion to generate touch signals. However, our conditioning comes from another touch signal rather than from a visual signal. Recent work aims to learn embeddings that work for multiple touch sensors~\cite{yang2024binding}. However, it is not generally possible to run off-the-shelf models on these representations, and they lack paired data to convert from one sensor to another. When it comes to tactile representations over a variety of sensors, recent work ~\cite{t3} presents an approach using unaligned tactile data from predominantly gel-based sensors. In contrast to this work, we leverage aligned touch across significantly different sensors, Soft Bubble (not gel-based) and GelSlim (gel-based), and demonstrate improved performance in downstream tasks. To the best of our knowledge, this work is the first to address cross-sensor touch-to-touch generation using aligned data.

\section{METHOD} 
\label{sec:methodology}

Our method has three main components. First, we use a robot to collect a dataset of paired data from two different touch sensors (Sec.~\ref{sec:capturing_setup}). Second, we train a cross-modal diffusion model to translate from one of these sensors to another (Sec.~\ref{sec:transfer_module}). Third, we evaluate the cross-modal prediction model using an in-hand pose estimation algorithm and use it to perform downstream tasks (Sec.~\ref{sec:evaluation_method}).

\begin{figure}
    \centering
    \includegraphics[width=\columnwidth]{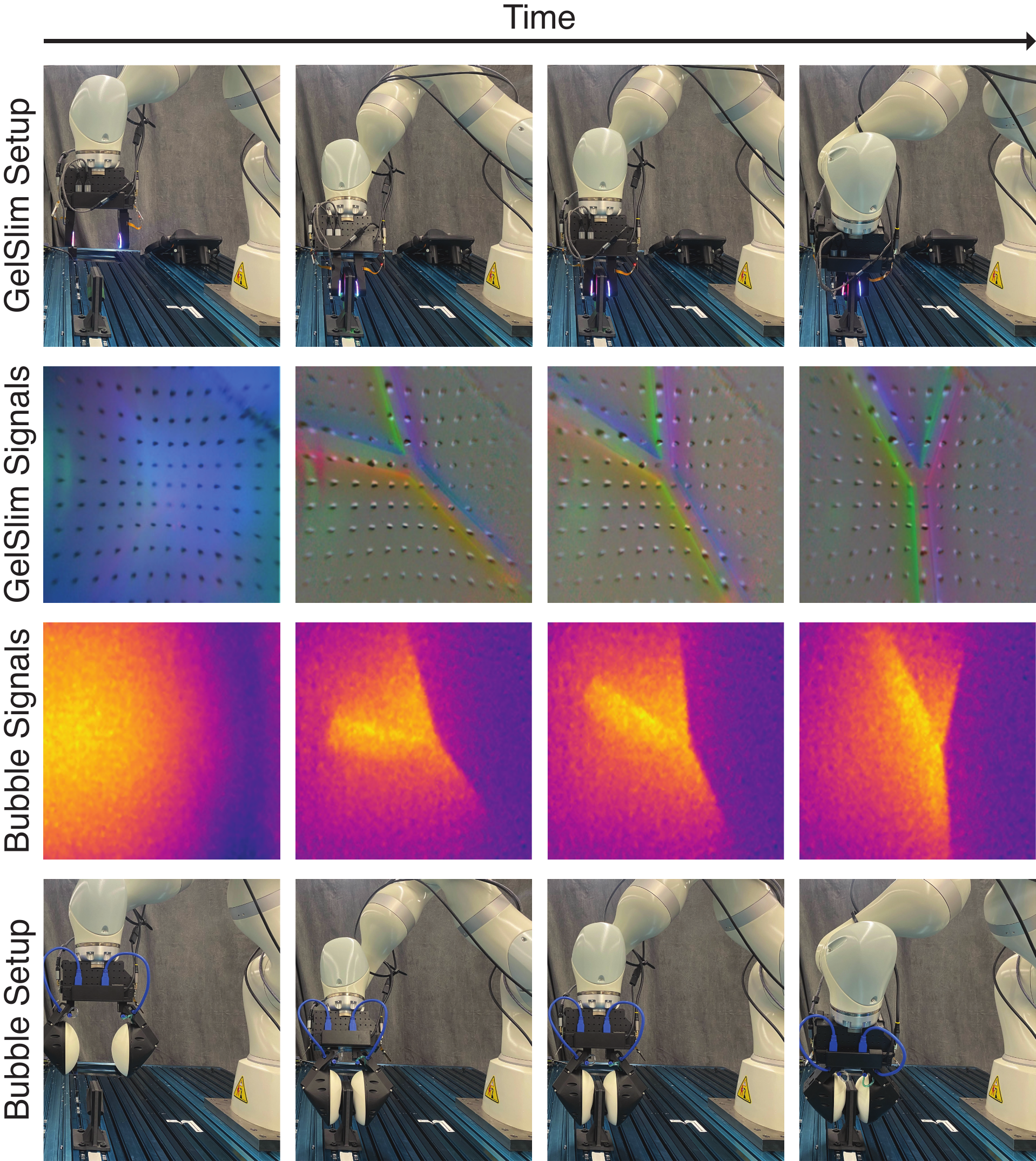}
    \caption{{\bf Collecting a dataset of paired touch signals.} To obtain paired touch data, we have a robot probe an object at the same position using two different touch sensors.}
    \label{fig:data_collect}
    \vspace{-0.6cm}
\end{figure}

\begin{figure}
    \centering
    \includegraphics[width=0.75\columnwidth]{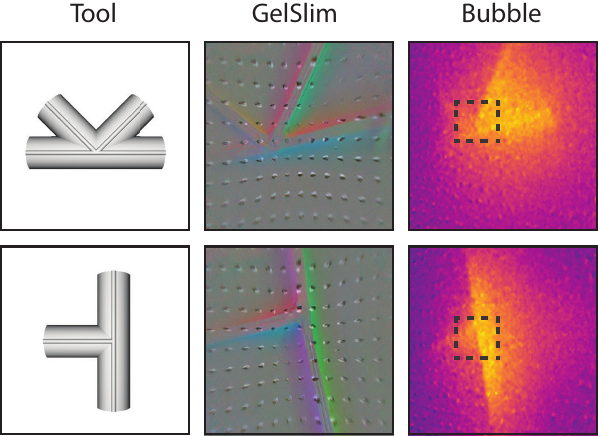}
    \caption{{\bf GelSlim and Soft Bubble touch signals.} We show 2 of the 12 tools from our dataset and the corresponding GelSlim and Soft Bubble images. The dashed rectangle over the Soft Bubble image indicates the (much smaller) contact area covered by GelSlim.}
    \label{fig:touch_signals}
    \vspace{-0.6cm}
\end{figure}

\subsection{Capturing Paired Multimodal Tactile Signals}
\label{sec:capturing_setup}
Our goal is to generate tactile signals that convey key information for downstream manipulation tasks like insertion and stacking. Therefore, we collect paired tactile signals for the Soft Bubble and GelSlim sensors using a setup that resembles a manipulation task. In our experiments we use a KUKA LBR Med 14 R820 robotic arm equipped with a WSG-50 gripper, as shown in Fig.~\ref{fig:data_collect}.
Given that our chosen sensors present significant differences in mechanical design, deformation at the contact interface, dimension of the contact area, and image resolution, we require a repeatable setup for pairing tactile signals that is capable of spatially precise and repeatable touches with either sensor.

To accomplish this, we ensure the Soft Bubble and GelSlim sensors are aligned by: First, the gripper is positioned such that the center of the sensors makes contact with a given point on the target object. To account for the greater compliance of the Soft Bubble sensors, the gripper is closed an additional 10mm after first contact. Because the GelSlim sensors deform significantly less, the gripper is only closed an additional 1mm after first contact. This procedure is critical to identify characteristic features of the target object with the highly compliant Soft Bubble sensors.

It is important that key features of objects are present in both sensor images to address the significant difference in contact area and tactile signal resolution between the Soft Bubble and GelSlim sensors. On one hand, the Soft Bubble sensor covers approximately $16 \times$ the contact area  of the GelSlim sensor, as shown in Fig.~\ref{fig:touch_signals}. On the other hand, the GelSlim sensor signals are much more detailed than the Soft Bubble sensor signals: they provide 23.72 pixels/mm versus 2.36 pixels/mm for the Soft Bubble. This difference in resolution allows the GelSlim to render precise microgeometry in the contact area.
To account for these disparities, we preserve key features of the tactile signatures across sensors by ensuring that all the touch signals in our dataset keep the distinctive features of each object (e.g., the elbow of a hex key) and by selecting objects that possess distinct features visible with the resolution of each sensor.

To demonstrate the importance of collecting images containing characteristic features of the tools, we also collected a set of paired images of potentially ambiguous images. Consider a GelSlim sensor placed so only a single straight shaft is visible. The true tool could be three of the four tools shown in in Fig.~\ref{fig:data_collect}. Due to the larger contact area of the Soft Bubble sensor, the generative model must infer which of the tools is in contact with the sensor and produce the geometry not captured in the GelSlim signal. Without distinguishing features visible in the GelSlim image, we should not expect the generative model to produce these features correctly. Sec.~\ref{sec:exp} reports empirical results confirming this hypothesis, where we examine the model's performance when exposed to ambiguous samples.
\subsection{A Dataset of Paired Touch Signals}
\label{dataset}
Using the above collection procedure, we obtain paired tactile signals for 12 different tools, two of which are shown in Fig.~\ref{fig:touch_signals}. The touch samples were collected within a 10mm x 10mm grid centered at the tool origin, with sensor angles in the range $\pm 22.5^{\circ}$. We collected 2,688 paired samples per tool for a total of approximately 32,256 paired samples. The dataset is split as follows: 19,350 for training, 6,453 for validation and 6,453 for testing. Details about the dataset composition are shown in the \href{https://www.mmintlab.com/research/touch2touch/}{project page}.

\subsection{Generating Touch from Touch}
\label{sec:transfer_module}

\begin{figure*}[!h]
    \centering
    \includegraphics[width=1\linewidth]{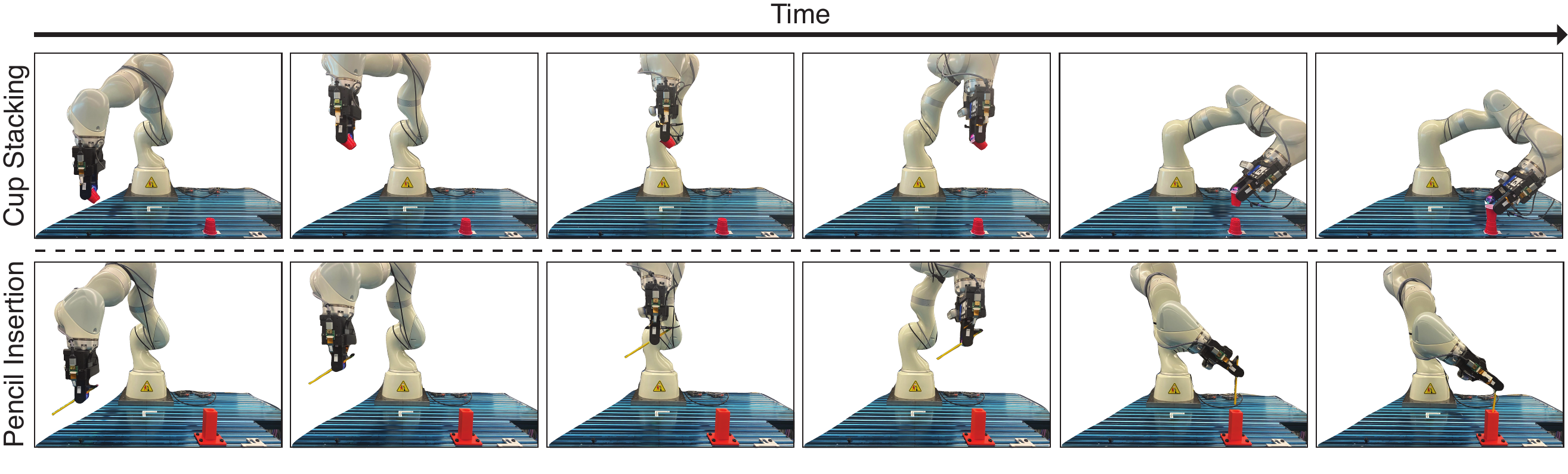}
    \vspace{-10mm}
    \caption{{\bf Downstream object manipulation tasks}. A robot arm is equipped with the GelSlim sensor.  It uses our model to estimate a corresponding Soft Bubble signal. Using this signal, it successfully completes stacking and insertion tasks, using an algorithm that operates on the Soft Bubble signal.}
    \label{fig:real_experiments}
    \vspace{-0.3cm}
\end{figure*}

\begin{figure}[h!]
    \centering
    \includegraphics[width=\columnwidth]{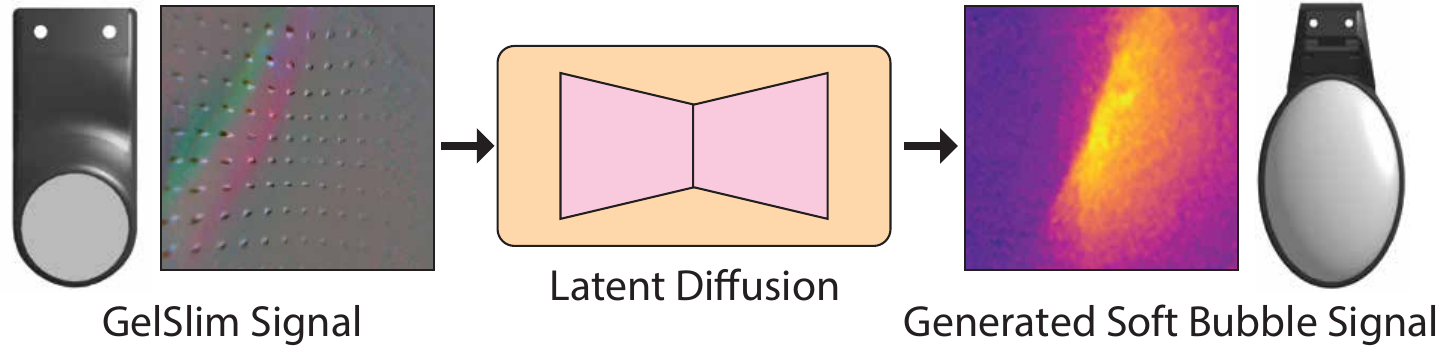}
    \vspace{-5mm}
    \caption{{\bf Cross-modal tactile generation}. We predict a Soft Bubble signal from a GelSlim signal using a diffusion model. To provide a conditioning signal, GelSlim images are encoded into 2D feature maps using a ResNet and concatenated channel-wise with the latent code. %
    }
    \label{fig:method}
    \vspace{-0.6cm}
\end{figure}

We use a generative model based on latent diffusion~\cite{rombach2022high} to estimate the Soft Bubble signals based on the GelSlim signals (Fig.~\ref{fig:method}). %
We use a ResNet-50 to encode the GelSlim image into a 2D feature map. The feature map is then concatenated with the noise and  passed into the denoising UNet. %
We subtract the empty background image from the GelSlim to reduce the possible effect of changing the gels or the sensor. The background image is captured by the GelSlim when it is not in contact with anything. %
We represent the Soft Bubble signal as a depth map, which we inflate into a 3-channel signal (by copying the image channel-wise 3 times). We train our model using random rotation, random flipping, and color jittering as augmentations.

Once the Soft Bubble tactile signal is generated from the GelSlim image using diffusion, we perform three post-processing steps to ensure accurate tactile information generation (beyond visual fidelity). First, we take the average value of the three channels of the prediction to map it back to one channel. Next, we need to normalize the prediction (in the range $(-1, 1)$) back to the depth map values by using the maximum and minimum values of the depth maps across the training dataset. Finally, to deal with small scaling/bias in the predictions (which can drastically change their interpretations as point clouds), we shift the pixel values of the generated Soft Bubble images. We calculate the mean and standard deviation of the generated and ground truth Soft Bubble images on the training dataset and use these values to renormalize the generated images. %

\subsection{In-Hand Pose Estimation and Downstream Tasks}
\label{sec:evaluation_method}
One of the ways that we evaluate our cross-modal touch prediction model is by estimating object pose from generated touch signals for downstream tasks.

\mypar{In-Hand Pose Estimation.}
\label{pose_estimation} To obtain object pose, we align the generated and measured Soft Bubble point clouds to their corresponding object geometry point clouds using ICP, which finds a rigid transformation between the two.  %
To perform ICP, we first need to obtain points belonging to the object surface from the full Soft Bubble point cloud.
We obtain these points by training a UNet ~\cite{unet} to generate a mask from the raw Soft Bubble images. Details of this method are included in the \href{https://www.mmintlab.com/research/touch2touch/}{project page}.
To acquire our second point cloud for ICP, we sample object surface points from a CAD model. The CAD model is aligned to the grasp frame of the sensor. We apply ICP to these point clouds to find the transformation that aligns the two and provides the relative pose of the Soft Bubble sensor with respect to the object. Using the inverse of this transformation, we obtain the in-hand pose estimate of the object with respect to the grasp frame of the sensor.

\mypar{Downstream Task: Peg-in-Hole Insertion.}
The peg-in-hole insertion task shown in Fig.~\ref{fig:real_experiments} consists of several steps: (1) we hand an object to the robot in a random orientation, (2) the robot grasps the object with GelSlim sensors, (3) we apply our cross-modal tactile generation model online to obtain a predicted Soft Bubble signal, (4) we perform ICP on the predicted Soft Bubble signal to estimate the angle of the object with respect to the grasp frame of the GelSlim sensor, (5) we align the object with the insertion hole using the estimated angle, and (6) we insert the object in the hole. 
We evaluate this task on three tools that were unseen during training and one real object (a pencil).

\mypar{Downstream Task: Cup Stacking.}
The stacking task utilizes a procedure similar to the insertion task. The main differences are the success criteria and the object geometry. The stacking task is completed successfully if one small SOLO-brand cup is balanced on top of another.

\section{EXPERIMENTS}
\label{sec:exp}
\begin{figure*}[t]
    \centering
    \includegraphics[width=1\linewidth]{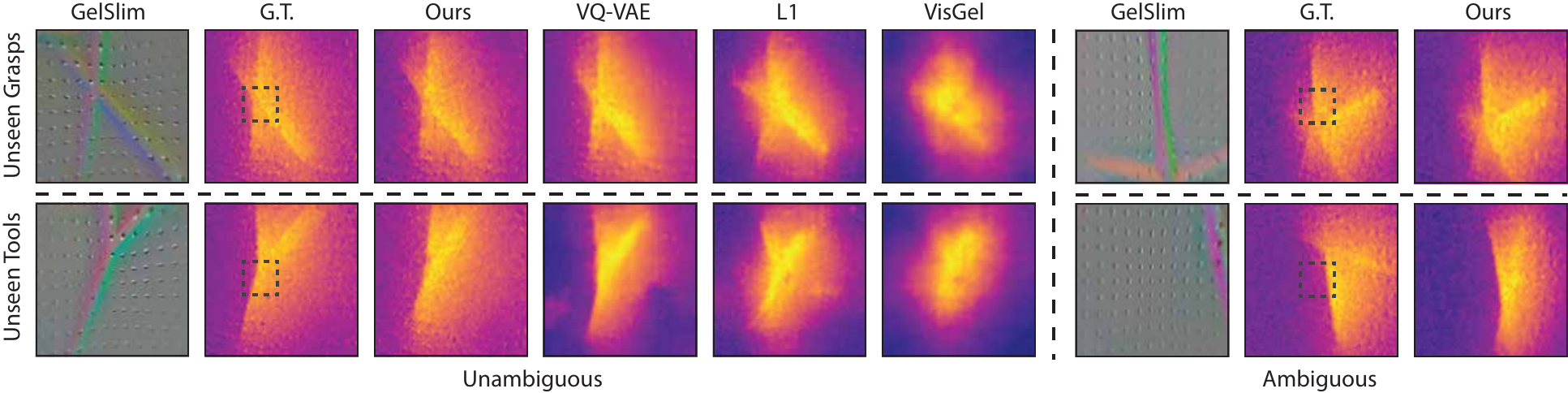}
    \vspace{-0.5cm}
    \caption{\textbf{Qualitative results for cross-modal touch prediction.} We show the generated bubble images tested on unseen grasps and unseen tools and compares them to baseline models used for image generation. In addition, we show results using our model architecture but trained and tested on an ambiguous dataset. We indicate the approximate size of the GelSlim sensor (which has a much smaller field of view) with a dashed gray box. }
    \label{fig:qualitative_results} \vspace{-0.5cm}
\end{figure*} 

\begin{figure}[h!]
    \centering
    \includegraphics[width=\columnwidth]{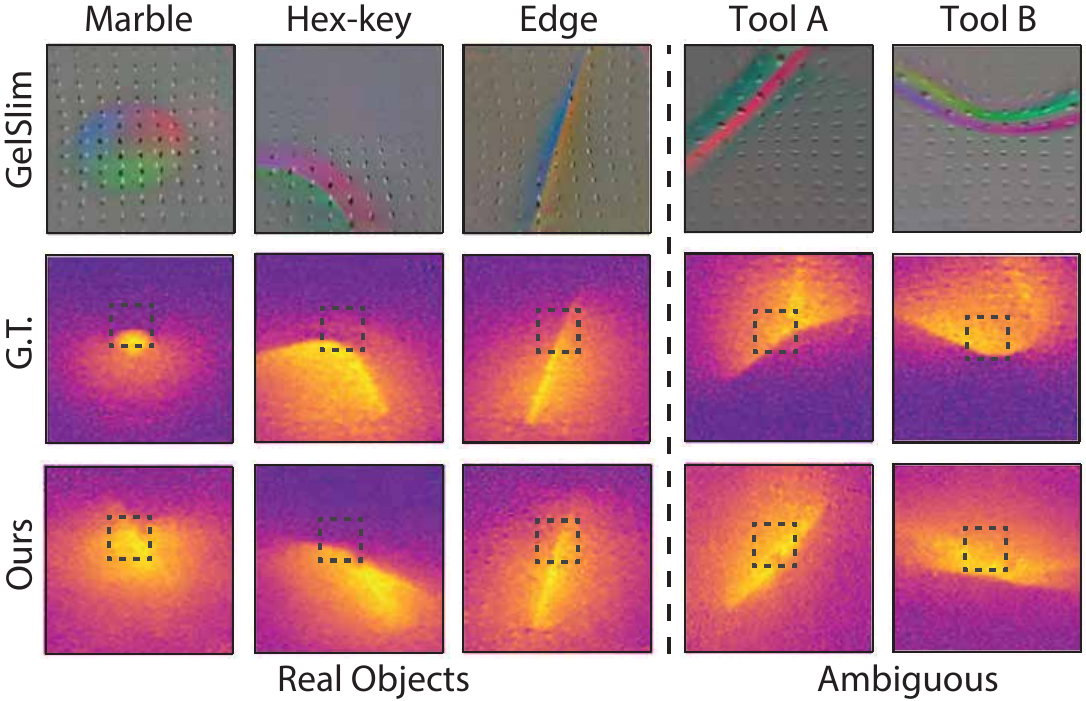}
    \vspace{-0.3cm}
    \caption{\textbf{Qualitative results for cross-modal touch prediction on real objects and ambiguous tools.} We indicate the approximate size of the GelSlim sensor (a much smaller field of view) with a dashed gray box. }
    \label{fig:qualitative_results_real_obj} \vspace{-0.8cm}
\end{figure}

\mypar{Experimental setup.}
We now evaluate the diffusion model's ability to generate Soft Bubble signals from GelSlim signals. We test this model for generalization to unseen grasps and to unseen tools. We select 3 tools to be fully unseen during training for the unseen tools evaluation dataset. 

\mypar{Evaluation metrics.}
We evaluate our method using two types of metrics: visual and functional metrics. \textit{Visual metrics} use standard image generation metrics to compare the ground truth signals from the Soft Bubble with our model's generated signals. Our selected visual metrics are Peak Signal to Noise Ratio (PSNR), Structural Similarity (SSIM), and Frechet Inception Distance (FID). These metrics are widely used in touch generation work~\cite{dou2024tactile,yang2023generating}. PSNR and SSIM are highly sensitive to spatial position. To evaluate our model beyond pixel-wise precision, we use FID. FID is a standard evaluation metric for cross-modal generation that measures the distance between the generated and ground-truth data distribution.
\textit{Functional metrics} test the suitability of our model's generated Soft Bubble signals for off-the-shelf ICP. ICP was selected due to its common use for estimating the relative pose of the Soft Bubble sensors with respect to grasped objects \cite{softbub_tedrake}.
To evaluate our method, we measure the error between the sensor angle estimated with ICP on the generated signal and the ground truth sensor angle obtained from robot proprioception.
 We calculated this error across our unseen grasps and unseen tools datasets. We also consider two thresholds for in-hand pose estimation success: ICP angle errors of ess than 5$^{\circ}$ or less than 10$^{\circ}$ from generated Soft Bubble signals. As an upper bound, we present the mean angle error, 5$^{\circ}$ success rate, and 10$^{\circ}$ success rate found when using the ground truth Soft Bubble signals to perform ICP. Finally, we compare our method to 3 baselines: a VQ-VAE \cite{van2017neural} with a fully convolutional architecture, VisGel \cite{li2019connecting}, and a UNet trained using L1 loss. We provide the implementation details of these models in the \href{https://www.mmintlab.com/research/touch2touch/}{project page}.

\mypar{Results.}
We evaluate the cross-modal tactile generation qualitatively and quantitatively, and apply it to downstream robotic tasks. Specifically, we show how dataset collection and data augmentation affect model performance and the importance of shifting the pixel values of the generated images as part of the post-processing. In addition, we compare our model with other generative models for cross-modal tactile generation. Finally, we demonstrate our model's potential to perform downstream robotic tasks designed for the Soft Bubble sensors using GelSlim sensors. 

Table ~\ref{tab:different_data_results} show the visual and functional metric results for different training datasets, respectively. The unambiguous dataset was collected using the procedure described in Sec.~\ref{sec:capturing_setup}. This dataset focuses on preserving key features of the grasped object on paired sensor signals. In contrast, the ambiguous dataset contains samples in which the GelSlim signal does not contain the same key features as the paired Soft Bubble signal. The \href{https://www.mmintlab.com/research/touch2touch/}{project page} contains details about the ambiguous dataset. In these tables, "aligned" refers to the paired tactile images that were aligned based on our procedure. Misalignment simulates an alignment error within 8 mm. Finally, we present a dataset where Gaussian noise was added to the conditioning images. In Table ~\ref{tab:different_data_results}, we can see that going from a fully unambiguous dataset to a mixed dataset and to a fully ambiguous dataset, the performance reduces significantly on the ICP metrics. In addition, when misalignment within 8 mm is introduced, the success is below 6\(\%\) for the 5$^{\circ}$. Regarding the dataset with noise, we observe a similar performance to our dataset. This is expected since the diffusion model training is based on denoising.

\begin{table}[h!]
\small
    \caption{Cross-modal generation performance on different tactile datasets evaluated on unseen tools. Amb., Unamb., Mixed, Al. and AE abbreviate Ambiguous, Unambiguous, Ambiguous + Unambiguous, Aligned and Angle Error.}
    \vspace{-0.3cm}
    \centering
    \renewcommand{\arraystretch}{0.4}
    \begin{tabularx}{\columnwidth}{X|C{0.4cm}C{0.4cm}C{0.6cm}|C{0.9cm}cc}
        \toprule
       Dataset  &  PSNR  & SSIM & FID 
                & AE ($^{\circ}$) & \multicolumn{2}{C{1.6cm}}{Success (\%)} \\
                & & & 
                & & 5$^{\circ}$ & 10$^{\circ}$\\
                &$\uparrow$ &$\uparrow$ &$\downarrow$
                &$\downarrow$ & $\uparrow$ & $\uparrow$\\
        \toprule
       Ground Truth         & - & - & - & 2.4   & 88.4 &  97.0\\
       \midrule
       Unamb. + Al.       & \textbf{20.4}   &    \textbf{0.47}   &   81.7 & 6.4   & \textbf{59.4} &  79.0\\ & & & & & &\\
       
       Mixed + Al.  & 14.5   &    0.20   &   128.3 & 32.1  & 11.6 &  22.5\\ & & & & & &\\
       
       Amb. + Al. & 14.0   &    0.22   &   119.1 & 37.1  & 10.1 &  20.3\\ & & & & & &\\
      
       Unamb. + Misal.           & 14.9   &   0.26    &   113.0 & 45.9  & 5.8  &   10.9\\ & & & & & &\\
       
       Unamb. + Noise                & \textbf{20.4}   &    \textbf{0.47}   &  \textbf{79.6} & \textbf{6.1 }  & 56.5 &  \textbf{84.1}\\
       \bottomrule
    \end{tabularx}
    \label{tab:different_data_results}
    \vspace{-0.6cm}
\end{table}

\begin{table*}[h!]
\small
    \centering
    \renewcommand{\arraystretch}{0.9}
        \caption{Cross-modal tactile generation baseline comparison evaluated on unseen grasps and unseen tools using visual and functional metrics. Our method is Diffusion + Shift. Unseen denotes unseen GelSlim sensor.}
        \vspace{-0.3cm}
        \begin{tabularx}{\linewidth}{X|C{0.8cm}C{0.8cm}C{0.8cm}|C{1.25cm}C{0.4cm}C{0.4cm}|C{0.8cm}C{0.8cm}C{0.8cm}|C{1.25cm}C{0.4cm}C{0.4cm}}
        \toprule
        Method  & \multicolumn{6}{c|}{\hspace{1pt}\rule[-0.1em]{2.6cm}{0.5pt}Unseen Grasps\hspace{1pt}\rule[-0.1em]{2.6cm}{0.5pt}} & \multicolumn{6}{c}{\hspace{1pt}\rule[-0.1em]{2.6cm}{0.5pt}Unseen Tools\hspace{1pt}\rule[-0.1em]{2.6cm}{0.5pt}}\\
                & PSNR  & SSIM & FID 
                & Angle ($^{\circ}$) & \multicolumn{2}{C{1.6cm}|}{Success (\%)} 
                & PSNR & SSIM & FID
                & Angle ($^{\circ}$) & \multicolumn{2}{C{1.6cm}}{Success (\%)}\\
                & & & 
                & Error & 5$^{\circ}$ & 10$^{\circ}$
                & & &
                & Error & 5$^{\circ}$ & 10$^{\circ}$\\
                &$\uparrow$ &$\uparrow$ &$\downarrow$
                &$\downarrow$ & $\uparrow$ & $\uparrow$ 
                &$\uparrow$ &$\uparrow$ &$\downarrow$
                &$\downarrow$ & $\uparrow$ & $\uparrow$\\
        \toprule
         Ground Truth        &   -             &     -     &     -    
                             & 0.96            & 98.6      &  100.0
                             &   -             &     -     &     -
                             & 2.4             & 88.4      &  97.0\\
        \midrule
         VisGel     & 20.4           &    0.30   &   179.5  
                    &  13.1          & 30.2      &  52.9
                    & 18.9           &    0.27   &   206.6
                    & 19.2           & 22.5      &  44.2\\
         L1         & 20.5           &    0.36   &   124.0  
                    &  9.4           & 43.2      &  63.8
                    & 19.1           &    0.32   &   156.0
                    & 15.9           & 24.6      &  47.1\\
         VQ-VAE     & \textbf{27.0}  &    0.57   &   144.2  
                    &  1.4           & 97.10     &  \textbf{100.0}
                    & \textbf{20.7}  &    0.37   &   212.8
                    & 8.4            & 40.6      &  73.2\\
         Diffusion        & 2.1     &    0.13   &   \textbf{60.8}   
                          &  28.1   & 14.7      &  28.7 
                          & 2.2     &   0.13    &   \textbf{81.7}
                          & 52.9    & 5.1       &   8.7\\ 
         \textbf{Ours} & 26.1           &    \textbf{0.62}   &   61.6
                                     & \textbf{1.3}   & \textbf{97.8}      &  99.3
                                     & 20.4           &    \textbf{0.47}   &  81.7
                                     & \textbf{6.4}   & \textbf{59.4}      & \textbf{79.0}\\
         \midrule
        Ours (Unseen) & 21.1   &    0.46   &  75.0 
                                         & 3.3    & 84.8      &  92.5
                                         & 19.6   &    0.38   &  93.8
                                         & 7.4    & 55.1      &  78.3\\
       
        \bottomrule
        \end{tabularx}
    \label{tab:quantitative_results} \vspace{-0.5cm}
\end{table*}

Table ~\ref{tab:quantitative_results} show that by shifting the pixel values (Sec.~\ref{sec:methodology}), our model significantly improves on PSNR, SSIM, and all the functional ICP metrics. However, the FID metric ~\cite{heusel2017gans} stays the same. An explanation for this could be that the perception of the images based on human judgment with and without shifting is almost the same. However, the latent diffusion model causes a distribution shift in the pixel values when generating the Soft Bubble images. This pixel shift would be negligible in terms of pure image generation evaluation. In contrast, we require higher precision with the pixel values when we want to use the generated tactile signatures for a downstream task.

Fig. \ref{fig:qualitative_results_real_obj} shows the qualitative results of our method being tested on real objects and ambiguous samples from two of our tools. For real objects, we noticed that the model performs better on line-like objects, especially in the region where the Gelslim is in contact (shown as a dotted rectangle). The model generates a blurred image with higher intensity within the dotted rectangle for circular-shaped objects. Regarding the ambiguous samples, we can observe how the model generates a tactile signal closer to ground truth inside the dotted rectangle and then out-paints a tool seen during training outside this rectangle.

We compare our method to other generative models in Table ~\ref{tab:quantitative_results}. Overall, diffusion and VQ-VAE perform similarly on the lower-level metrics (PSNR and SSIM). This is partly because the VQ-VAE is directly optimized on the L1 loss, which is highly correlated with the low-level, pixel-wise metrics. On FID, which is a higher-level metric, the diffusion model performs notably better, indicating that it is more capable of capturing the distribution of the Soft Bubble images. We also evaluate the model performance using the functional metrics. On these metrics, the diffusion model achieves much higher accuracy on the testing data of both unseen grasps and unseen tools, indicating that the predicted images are not only visually accurate but also have the potential to be applied to robotic tasks that require dense spatial information. We show qualitative results in Fig.~\ref{fig:qualitative_results}.

To test our diffusion model on another downstream task designed for Soft Bubble images, we trained a tool classification model on Soft Bubble images and zero-shot evaluate the model on the generated Soft Bubble images from GelSlim signals. We obtain 88.1\% accuracy when evaluated on real Soft Bubble images and 78.7\% when evaluated on generated Soft Bubble images. These results show that we can use our model to use a tool classification model previously trained on real Soft Bubble images with GelSlim sensors.

We test our diffusion model on different GelSlim sensors and show in Table \ref{tab:quantitative_results} that the drop in accuracy for unseen tools is below 5\(\%\) for the 5$^{\circ}$ threshold and below 1\(\%\) for the 10$^{\circ}$ threshold. This shows that our model is robust to unseen GelSlims with slightly different colors and dot orientations.

\begin{table}[!]
\small
    \centering
    \caption{Ablation study on augmentation techniques showing the impact of applying random cropping, rotation, flipping, and padding. The full model is shown in the final row.}
    \vspace{-0.3cm}
    \begin{tabularx}{\columnwidth}{X c c c c}
        \toprule
            Method & Angle $\downarrow$ & \multicolumn{2}{c}{Success (\%) $\uparrow$} \\
           & Error ($^{\circ}$) & 5$^{\circ}$ & 10$^{\circ}$\\
        \midrule
            Ground Truth                    &2.4  & 88.4  & 97.1 \\
            Diffusion + Shifting            &6.6  & 55.1  &79.7\\
        ~~~ +Padding               &30.8 & 13.0  & 18.1\\
        ~~~ +Padding \& Cropping   &37.4 & 6.6   & 11.6 \\
        ~~~ +Rotation              &8.0  & \textbf{59.4}  & 73.9\\
        ~~~ +Rotation \& Flipping   &\textbf{6.2}  & 58.7  & \textbf{83.3} \\
        \bottomrule
    \end{tabularx}
    \label{tab:ablation} \vspace{-0.3cm}
\end{table}

\vspace{-0.2cm}
\begin{table}[h!] %
\small
    \centering
    \renewcommand{\arraystretch}{0.9}
    \caption{Insertion and Cup Staking Tasks Success on Unseen Objects}
    \vspace{-0.3cm}
    \begin{tabularx}{0.75\columnwidth}{X c c}
        \toprule
            Tool & \multicolumn{2}{c}{Success Rate} \\ %
            & Diffusion & VQ-VAE \\
        \toprule
            Tool 1 Insertion & 18/30 & 9/30\\
            Tool 2 Insertion & 10/30 & 8/30\\
            Tool 3 Insertion & 15/30 & 15/30\\
        \midrule
            Pencil Insertion & 21/30 & 7/30\\
            Cup Stacking & 22/30 & 21/30\\
        \bottomrule
    \end{tabularx}
    \label{tab:Insertion Task Success} \vspace{-0.3cm}
\end{table}

\mypar{Ablation study.}
Table \ref{tab:ablation} shows that adding random rotations and flipping to our model improves our performance on unseen tools, while padding causes a drop in performance. We attribute this drop in performance to the loss of tactile signal from the GelSlim sensor. In this case, padding refers to surrounding the GelSlim image with zeros such that its spatial location matches the Soft Bubble image. When the GelSlim images are padded, they occupy only 1/16 of the pixels in the full padded image. We can see this in Fig.~\ref{fig:qualitative_results}: the GelSlim image only corresponds to a small portion of the Soft Bubble image.  Even though padding the image may help facilitate the alignment of tactile signatures between the Soft Bubbles and GelSlim images, it offers sparse tactile information to condition our diffusion model. This negatively affects the performance of the cross-modal generation model.

\mypar{Downstream task performance.}
For downstream task performance, we show the results in Table \ref{tab:Insertion Task Success}. For the insertion and stacking tasks, we show the success rate across 30 trials with diffusion and VQ-VAE models. In general, they perform similarly. However, diffusion performs significantly better on Tool 1 and pencil for insertion. Overall, diffusion shows a 57.33$\%$ success rate and VQ-VAE a 40.00$\%$ success rate. These values are close to the success rate at the 5$^{\circ}$ threshold for unseen tools shown in Table  \ref{tab:quantitative_results} for these models, which is closely related to the downstream task. %

\vspace{-0.3cm}
\section{Discussion}

In this paper, we have proposed a method for predicting the signal of one touch sensor from another and applied our model to object manipulation tasks. %
Our work opens three possible directions for future research. First, it opens the possibility of transferring other models between touch sensors, which in the past have often required sensor-specific methods. Second, it opens research in new models for cross-modal touch translation. Finally, we anticipate that cross-modal touch prediction will improve our understanding of mutual information content across tactile sensors.

\mypar{Limitations.}
While our technical approach does not make explicit assumptions on the design of the sensor or the image representation, our experimental results focused on two vision-based touch sensors: the Soft Bubble and GelSlim. We chose these sensors since they have large differences between them, but other pairs of sensors may have unique challenges. Our diffusion-based approach assumes that the underlying signal is an image, a common assumption in vision-based touch sensors that are ubiquitous in robotics. However, this assumption may not be applicable to all touch sensors, especially touch sensors that do not directly perceive shape (e.g., BioTac). Regarding these sensors' interaction with objects, our method does not directly address differences in intrinsic contact dynamics between the Soft Bubble and the GelSlim, which could potentially be necessary when implementing cross-modal tactile generation on more contact-rich tasks.

\addtolength{\textheight}{-12cm}   %

\bibliographystyle{IEEEtran}
\bibliography{ref}

\end{document}